\documentclass[letterpaper, 10 pt, journal, twoside]{IEEEtran}

\usepackage{url}
\usepackage{wrapfig}
\usepackage{graphicx}
\usepackage{caption}
\usepackage{amsmath}
\usepackage{amssymb}
\usepackage{multirow}
\usepackage{booktabs}
\usepackage{color}
\usepackage{xcolor}
\usepackage{cite}
\definecolor{citecolor}{HTML}{0071bc}

\usepackage[pagebackref=false,breaklinks=true,letterpaper=true,urlcolor=citecolor,colorlinks,citecolor=citecolor,bookmarks=false]{hyperref}

\usepackage[export]{adjustbox}

\newlength\savewidth\newcommand\shline{\noalign{\global\savewidth\arrayrulewidth
  \global\arrayrulewidth 1pt}\hline\noalign{\global\arrayrulewidth\savewidth}}
\newcommand{\tablestyle}[2]{\setlength{\tabcolsep}{#1}\renewcommand{\arraystretch}{#2}\centering\footnotesize}

\definecolor{demphcolor}{RGB}{144,144,144}
\newcommand{\demph}[1]{\textcolor{demphcolor}{#1}}

\newcommand\blfootnote[1]{%
  \begingroup
  \renewcommand\thefootnote{}\footnote{#1}%
  \addtocounter{footnote}{-1}%
  \endgroup
}

\begin{document}

\title{Learning Continuous Grasping Function with \\ a Dexterous Hand from Human Demonstrations}

\author{Jianglong Ye$^{1*}$, Jiashun Wang$^{2*}$, Binghao Huang$^1$, Yuzhe Qin$^1$, Xiaolong Wang$^1$%
}

\markboth{IEEE Robotics and Automation Letters. Preprint Version. Accepted March, 2023}
{Ye \MakeLowercase{\textit{et al.}}: Learning Continuous Grasping Function with a Dexterous Hand from Human Demonstrations}

\twocolumn[{%
\renewcommand\twocolumn[1][]{#1}%
\maketitle
\begin{center}
    \vspace{-0.15in}
    \centering
    \captionsetup{type=figure}
    \includegraphics[width=0.95\linewidth]{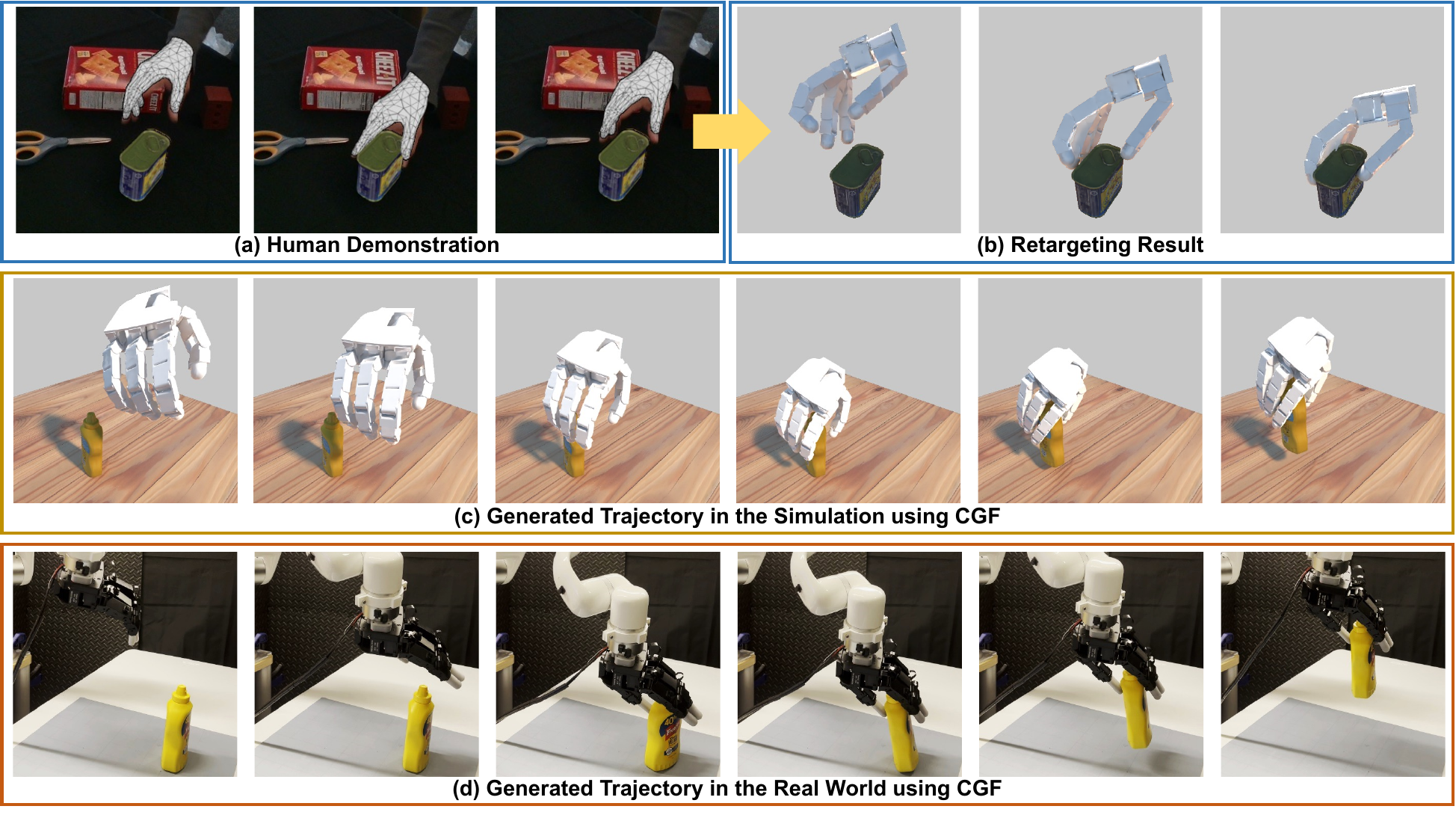}
    \vspace{-0.1in}
    \captionof{figure}{\small Examples of our generated trajectories learned from human demonstrations. Given hand-object trajectories from human video (a), we first translate them into robot manipulation demonstrations (b). We then train Continuous Grasping Function (CGF) to generate human-like trajectories and deploy them in simulation (c) and real robot (d).}
    \vspace{-0.05in}
    \label{fig:teaser}
\end{center}
}]

\blfootnote{Manuscript received: November, 23, 2022; Revised February 08, 2023; Accepted March, 03, 2023.}
\blfootnote{The first two authors contributed equally. Jianglong Ye, Binghao Huang, Yuzhe Qin and Xiaolong Wang are with $^1$University of California San Diego. Jiashun Wang is with Carnegie Mellon University. Correspondence at {\tt\footnotesize xiw012@ucsd.edu}.}
\blfootnote{This paper was recommended for publication by Editor Hong Liu upon evaluation of the Associate Editor and Reviewers' comments.}
\blfootnote{This project was supported, in part, by NSF CCF-2112665 (TILOS), NSF CAREER Award IIS-2240014, NSF 1730158 CI-New: Cognitive Hardware and Software Ecosystem Community Infrastructure (CHASE-CI), NSF ACI-1541349 CC*DNI Pacific Research Platform, the Industrial Technology Innovation Program (20018112, Development of autonomous manipulation and gripping technology using imitation learning based on visualtactile sensing) funded by the Ministry of Trade Industry and Energy of the Republic of Korea, Amazon Research Award and gifts from Qualcomm.}
\blfootnote{Digital Object Identifier (DOI): see top of this page.}

\vspace{-0.14in}
\begin{abstract}
We propose to learn to generate grasping motion for manipulation with a dexterous hand using implicit functions. With continuous time inputs, the model can generate a continuous and smooth grasping plan. We name the proposed model Continuous Grasping Function (CGF). CGF is learned via generative modeling with a Conditional Variational Autoencoder using 3D human demonstrations. We will first convert the large-scale human-object interaction trajectories to robot demonstrations via motion retargeting, and then use these demonstrations to train CGF. During inference, we perform sampling with CGF to generate different grasping plans in the simulator and select the successful ones to transfer to the real robot. By training on diverse human data, our CGF allows generalization to manipulate multiple objects. Compared to previous planning algorithms, CGF is more efficient and achieves significant improvement on success rate when transferred to grasping with the real Allegro Hand. Our project page is available at \url{https://jianglongye.com/cgf/}.
\end{abstract}


\IEEEpeerreviewmaketitle

\section{Introduction}
\label{sec:introduction}

\IEEEPARstart{L}{earning} to perform grasping with a multi-finger hand has been a long-standing problem in robotics~\cite{salisbury1982articulated,rus1999hand,okamura2000overview,Dogar2010}. Using a dexterous hand instead of a parallel gripper offers the robot the flexibility on operating with daily life objects like humans do, but also largely increases the difficulty given the large Degree-of-Freedom of the dexterous hand. A typical method for this task is a 2-step paradigm including grasp pose estimations following by motion planning~\cite{varley2015generating,brahmbhatt2019contactgrasp,lu2020planning}. Recent works have also studied on using Reinforcement Learning with human demonstration guidance for grasping~\cite{mandikal2021learning,qin2021dexmv}.

While these approaches have shown encouraging results, they plan the grasping with finite discrete time steps. On the other hand, human grasping motion is continuous, can we learn a continuous grasping process for robot hands? Making robot grasping continuous can lead to a more natural and human-like trajectory, and each step we sample will be differentiable which we can use a more robust PD control with feedforward. Recent progress on neural implicit functions have shown successful applications in learning continuous image representation~\cite{chen2021learning,dupont2021generative} and continuous 3D shape representation~\cite{park2019deepsdf,mescheder2019occupancy,mildenhall2020nerf}. Can this success be migrated from representing 2D/3D space to time?

In this paper, we propose to learn Continuous Grasping Function (CGF) with a dexterous robotic hand. To mimic the continuous human motion, we utilize human grasp trajectories from videos to provide demonstrations and supervision in training. By training CGF with generative modeling on a large-scale of human demonstrations, it allows generalization to grasp multiple objects with a real Allegro robot hand as shown in Figure~\ref{fig:teaser} (d).

Specifically, given the 3D hand-object trajectories from human videos (Figure~\ref{fig:teaser} (a)), we first perform motion retargeting to convert the human hand motion to the robotic hand motion to obtain the robotic manipulation demonstrations (Figure~\ref{fig:teaser} (b)). We then learn a CGF in the framework of a Conditional Variational AutoEncoder (CVAE)~\cite{Sohn2015LearningSO} by reconstructing the robotic hand motion with these demonstrations. Specifically, the conditional encoder of our CGF model will take the object point clouds as inputs and provides the object embedding. Taking the concatenation of the object embedding, a latent code $z$ and a time parameter $t$, the decoder of CGF is an implicit function which outputs the dexterous hand parameters in the corresponding time $t$. By sampling a continuous-time sequence of $t$, we can recover a continuous grasping trajectory in any temporal resolution. By sampling the latent code $z$, we can achieve diverse trajectories for the same object.  Figure~\ref{fig:teaser} (c) shows an example of the inferred grasping trajectory in the simulator.

In our experiments on testing our model, we will perform sampling on the latent code $z$ multiple times given a test object and generate diverse grasping trajectories. We then execute these trajectories in the simulator and select the one which can successfully grasp the object up. Different from the previous paradigm on grasping followed by planning, we empirically find our method is much more efficient since we avoid performing planning for each trajectory but directly generate the trajectory from CGF. Given the selected trajectories from the simulator, we can deploy them in the real world with an Allegro hand attached on an X-Arm 6 robot. Compared with planning, our method achieves better Sim2Real generalization with more natural and human-like motion, which leads to a better success rate.

We highlight our main contributions here: (i) A novel Continuous Grasping Function (CGF) model which allows smooth and dense sampling in time for generating grasping trajectory; (ii) CGF allows efficient generation of grasping plan and more robust control in simulation; (iii) We achieve much significant improvement on Sim2Real generation on Allegro hand by learning CGF from human demonstrations.

\vspace{-0.05in}
\section{Related Work}
\label{sec:related}

\textbf{Generalization in Dexterous Manipulation.} Dexterous manipulation is one of the most challenging problems in robotics~\cite{salisbury1982articulated,rus1999hand,okamura2000overview,Dogar2010,dafle2014extrinsic,calli2018path}. Recent studies on model-free~\cite{Openai2018,Openai2019,huang2021generalization,chen2022system} and model-based~\cite{kumar2016optimal,nagabandi2020deep} Reinforcement Learning (RL) have achieved encouraging results on multiple complex dexterous manipulation tasks.
However, there is still a large challenge on generalization for RL. For example, when the RL policy in~\cite{Openai2018} can be transferred to the real robot, it is learned specifically for one object. On the other hand, an RL policy trained with multiple objects in simulator~\cite{chen2022system} has not yet been transferred to real. Instead of using RL, one line of works on dexterous grasping is first performing a grasp estimation and then planning for execution~\cite{Andrews2013, varley2015generating, lu2020planning}, which have shown great success on generalization to multiple objects and in real robots at the same time.
Our work aligns more closely with this line of research. Through training on large-scale human demonstrations, our method is able to generate grasping trajectories for unseen objects based on their geometry.  Compared to previous planning methods, our approach provides more diverse, smooth and natural grasping trajectories in a more efficient way, as evidenced by our experimental results.

\textbf{Grasping Motion Synthesis.} Synthesizing and predicting human grasp has been an active research field for both computer vision and
robotics~\cite{DBLP:conf/rss/MorrisonLC18, brahmbhatt2019contactgrasp,karunratanakul2020grasping,jiang2021hand}.
For example, Grasping Field~\cite{karunratanakul2020grasping} is proposed as an implicit function that generates plausible human grasps given a 3D object. However, to apply on the robot hand, we will need to synthesize the full motion instead of a static pose. This motivates the research on synthesizing the hand-object interaction motions~\cite{hsiao2006imitation,ye2012synthesis,wu2021saga,taheri2022goal}.
For example, a full body and hand motion are synthesized together to grasp an object in~\cite{wu2021saga}. While related to our work, most approaches are still focusing on modeling the human hand. In this paper, we provide a framework where we first retarget the human hand trajectories to the robot hand trajectories and learn the robot grasping function with them.

\textbf{Learning from Human Demonstrations.} Our work is related to imitation learning or RL with human demonstrations for not only parallel grippers~\cite{schmeckpeper2020reinforcement,shao2020concept,young2020visual} but also dexterous hands~\cite{gupta2016learning,christen2019guided,qin2021dexmv,sivakumar2022robotic,qin2022one}.
For example, DexMV is a platform proposed in~\cite{qin2021dexmv} for extracting 3D human demonstrations from videos, generating robot demonstrations by retargeting, and augmenting Reinforcement Learning with these demonstrations for multiple manipulation tasks.
Although both methods utilize robot demonstrations for training, DexMV's use of GT states as inputs limits its generalizability to multiple objects and real robot transfer. In contrast, our implicit function generates continuous grasping given a point cloud input and can be deployed to the real robot. As most RL approaches are still with full access to GT states, they are not directly comparable to our method.

\begin{figure*}
\centering
  \vspace{-0.1in}
  \includegraphics[width=0.95\linewidth]{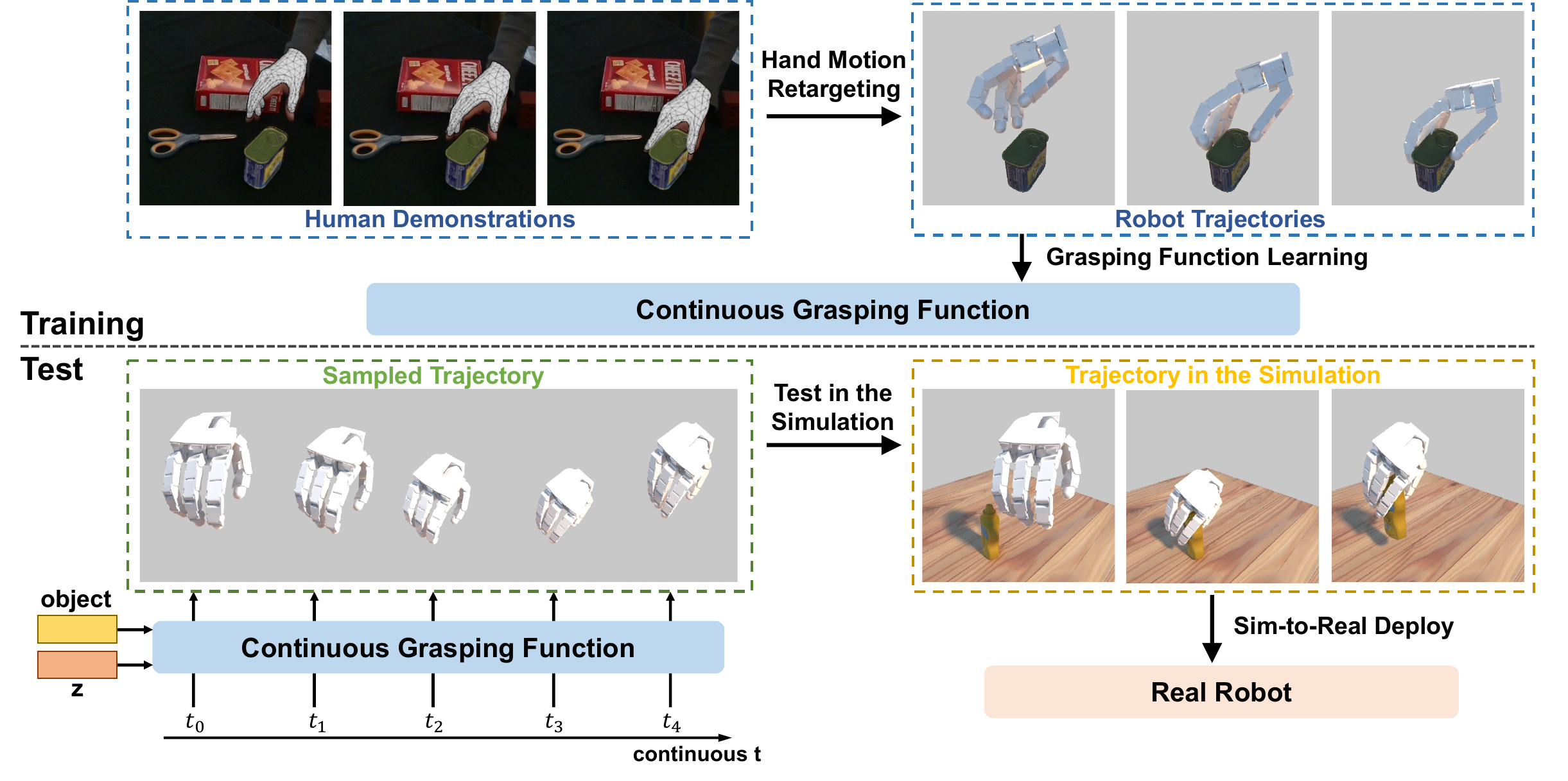}
  \vspace{-0.15in}
  \caption{\small
  Pipeline overview. During training, human demonstrations are first translated to robot joint positions which serve as the supervision for grasping function learning. During inference, our trained CGF takes a sampled latent code $z$, object feature and query time sequence as inputs to generate the trajectory. We then execute these trajectories in the simulator and deploy successful ones to the real robot.}
  \vspace{-0.15in}
  \label{fig:overview}
\end{figure*}

\textbf{Implicit Functions for Robotic Tasks.} Beyond its successful applications in computer vision, implicit functions have recently been explored in robotic manipulation tasks~\cite{li2021learning,simeonov2021neural,li20223d,jiang2021synergies,adamkiewicz2022vision}. For example, NeRF~\cite{mildenhall2020nerf} is used as a manner for learning 3D representations for control in~\cite{li2021learning}. Instead of using implicit functions to learn static 3D representations, our work focuses on the continuity in time. We build a grasping function to directly generate the trajectory instead of a static scene.

\section{Method}
\label{sec:method}

\vspace{-0.05in}
\subsection{Overview}

We aim to learn human-like robot grasping given object point clouds as input. We emphasize that learning from human demonstration could lead to more natural trajectories and the continuity helps the following control. To this end, we train Continuous Grasping Function with a CVAE framework to generate continuous trajectories and deploy them in the simulator and real robot consecutively. The pipeline is shown in Fig.~\ref{fig:overview}. During training, we first perform hand motion retargeting on a large-scale hand-object interaction dataset~\cite{chao2021dexycb} to collect demonstrations. Then the retargeting results served as the supervision for the grasping function learning. During inference, numerous continuous trajectories are sampled for a specific object and tested in the simulator. The successful trajectories will be deployed to the real robot. Besides, we can take more advantage of continuous implicit function by utilizing PD control with feedforward with the derivative of the joint positions.

\vspace{-0.05in}
\subsection{Human Demonstration Translation}
\label{subsec:retargeting}

Data collection on human hand-object interactions is relatively well established and accessible. Using large-scale human hand-object interaction data, we can learn patterns of how dexterous hands manipulate objects, and our goal is to generalize it to the robot hand. In this paper, we use ground truth trajectories from DexYCB~\cite{chao2021dexycb}. Translating human hand motion to robotics motion is the first step. Following~\cite{qin2022one}, we formulate our hand motion retargeting problem as a position-based optimization problem. We encourage the robot's joint position to be as close as possible to its corresponding human hand joint position,
\begin{equation}
\begin{aligned}
\min_{q_t}& \quad \sum^{N}_{i=0}||\mathcal{J}_i(q_t)-j_i||^2+\lambda ||\mathcal{J}_i(q_t)-\mathcal{J}_i(q_{t-1})||^2 \\
\text{s.t.}& \quad q_{lower} \leq q_t \leq q_{upper},
\end{aligned}
\end{equation}
where $q_t$ is joint position at time $t$, $\mathcal{J}_i$ is the forward kinematics function of the i-th joint and $j_i$ is the Cartesian coordinates of the hand joint which matches the i-th joint of the robot. $q_{lower}$ and $q_{upper}$ are the joint limits. We also encourage smoothness by incorporating a normalization term to penalize a large distance between $q_t$ and $q_{t-1}$. We set the initial $q_0=\frac{1}{2}(q_{lower}+q_{upper})$.

\vspace{-0.05in}
\subsection{Continuous Grasping Function Learning}

\begin{figure*}
\centering
  \vspace{-0.05in}
  \includegraphics[width=0.9\linewidth]{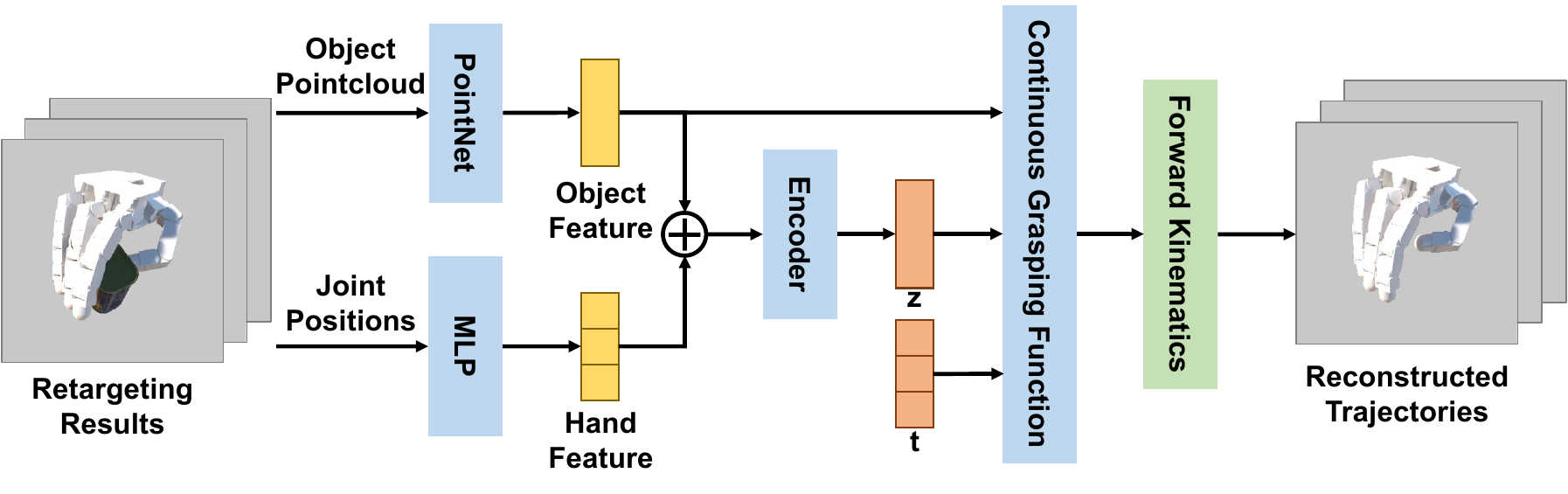}
  \vspace{-0.15in}
  \caption{\small
  Network architecture. Our generative model takes object point cloud and a sequence of joint positions as input and recovers corresponding robot hands. The proposed CGF takes the latent code $z$, object feature, and the query time $t$ as inputs to predict the corresponding joint position $\hat{q_t}$. $\oplus$ denotes concatenation. }
  \vspace{-0.2in}
  \label{fig:arch}
\end{figure*}

Our generative model is based on Conditional Variational Auto-Encoder (CVAE)~\cite{sohn2015learning}, where we propose Continuous Grasping Function (CGF) to replace the traditional decoder. During training, both encoder and CGF are used to learn the grasping generation in a robot hand reconstruction task with object point clouds and robot joint positions as inputs; during inference, only CGF is used to generate continuous grasping with only object point clouds as input. The architecture is shown in Figure~\ref{fig:arch}.

\textbf{During training}, given the object point cloud $\mathcal{P}^{o} \in \mathbb{R}^{N \times 3}$ ($N$ is the number of points) and a sequence of joints positions $\{ q_t \}, t \in \{0, 1, \cdots, T\}$ ($T$ is the number of frames) as inputs, we employ PointNet~\cite{qi2017pointnet} and MLP to extract object feature $\mathcal{F}^{o} \in \mathbb{R}^{1024}$ and hand features $\{\mathcal{F}^{h}_t\} \in \mathbb{R}^{256}$ respectively. All these features are then concatenated as $\mathcal{F}^{oh}$ for the encoder input. The outputs of the encoder are the mean $\mu \in \mathbb{R}^{256}$ and variance $\sigma^{2} \in \mathbb{R}^{256}$ of the Gaussian distribution $\mathcal{Q}\left(z \mid \mu, \sigma^{2}\right)$. The latent code $z$ is sampled from the distribution for the hand reconstruction.

After the encoding and sampling, we use our CGF to decode continuous grasping. Inspired by implicit functions~\cite{park2019deepsdf, mescheder2019occupancy, chen2019learning} for shape representation, our proposed CGF maps the query time $t$ to the joint position $q_t$. We concatenate latent code $z$ and the object feature $\mathcal{F}^{o}$ with time $t$ as the input for CGF. Specifically, CGF is a MLP $f$ parameterized by $\theta$ which predicts the corresponding joint position $\hat{q_t}$:
\begin{equation}
    \hat{q_t} = f(t, z, \mathcal{F}^{o}; \theta).
\end{equation}
We reverse the time and define $t=0$ as the end of the grasping for the convenience of implementation, i.e., when the robot's hand touches the object. And as $t$ grows, the hand moves further away from the object.  Given the predicted joint position $\hat{q_t}$, a differentiable forward kinematics layer $\mathcal{J}_i$ is utilized to get the Cartesian coordinate of the i-th joint.

The first objective function is the reconstruction error, which is defined as the $\mathcal{L}2$ distance between the joint positions as well as their Cartesian coordinates. We denote them as $\mathcal{L}_q =\sum_{t=0}^{T-1} \|\hat{q_t}-q_t\|_{2}^{2}$ and $\mathcal{L}_j = \sum_{t=0}^{T-1} \sum_{i=0} \|\mathcal{J}_i(\hat{q_t}) - \mathcal{J}_i(q_t)\|_{2}^{2}$ respectively. Following the training of VAE~\cite{Andrews2013}, we define a KL loss to encourage the latent code distribution $\mathcal{Q}\left(z \mid \mu, \sigma^{2}\right)$ to be close to the standard Gaussian distribution, which is achieved by maximizing the KL-Divergence as $\mathcal{L}_{KL} = -D_{KL}\left(Q\left(z \mid \mu, \sigma^{2}\right)\| \mathcal{N}(0, I)\right)$. We also introduce a contact loss at the end of the grasping to push the tips of robot hand to the object surface, which is achieved by minimizing distances to their closest object points $\mathcal{L}_{contact} = \sum_{i} \min_{p \in \mathcal{P}^{o}} \| \mathcal{J}_i(q_0) - p  \|^2_2$ where $i$ belongs to the indices of all tips. The full training loss is:
\begin{equation}
    \mathcal{L} = \lambda_q \mathcal{L}_q + \lambda_j \mathcal{L}_j + \lambda_{KL} \mathcal{L}_{KL} + \lambda_{c} \mathcal{L}_{contact},
\end{equation}
where $\lambda_q, \lambda_j, \lambda_{KL}$ and $\lambda_c$ are weights for various losses.

\textbf{During inference}, our CGF can easily sample a large number of diverse trajectories. We use PointNet to get the object feature $\mathcal{F}^o$ and sample a random latent code $z$ from the standard Gaussian distribution, then with a time query sequence, our CGF can produce a continuous and natural trajectory. By sampling a large amount of $z$, we can find trajectories with a smaller gap between the human hand and robot hand, which guarantees both natural and successful trajectories.

\vspace{-0.05in}
\subsection{PD Control with Feedforward}
\label{sec:pd-control}
Different from previous works, we utilize an implicit function to output the target joint position $q_d$, and given the continuity property of the implicit function, we can easily get the derivative (target joint velocity) $\dot{q}_d$ and the second-order derivative (target joint acceleration) $\ddot{q}_d$. Thus we can use a more robust controller, PD control with feedforward in the form of
\begin{equation}
    \tau = \text{ID}(\ddot{q}_d, q, \dot{q})-K_pe-K_v\dot{e},
\end{equation}
where $e=q-q_d$, $\dot{e}=\dot{q}-\dot{q}_d$, $\tau$ is the joint torque and $\text{ID}(\ddot{q}_d, q, \dot{q})$ is the inverse dynamics. $K_p$ and $K_v$ are hyperparameters. $q$, $\dot{q}$, and $\ddot{q}$ are the joint position, velocity, and acceleration of the robot. As far as we know, our method is the first end-to-end manner that can directly get the derivative to use the PD control with feedforward.

\vspace{-0.1in}
\section{Experiments}
\label{sec:experiments}

We conduct quantitative and qualitative evaluations in both simulator and real world. We show that by learning from human demonstrations, CGF is more efficient in finding successful grasping and our generation results can be transferred to the real robot with a higher success rate.

\begin{figure*}[t]
\centering
  \vspace{-0.1in}
  \includegraphics[width=0.85\linewidth]{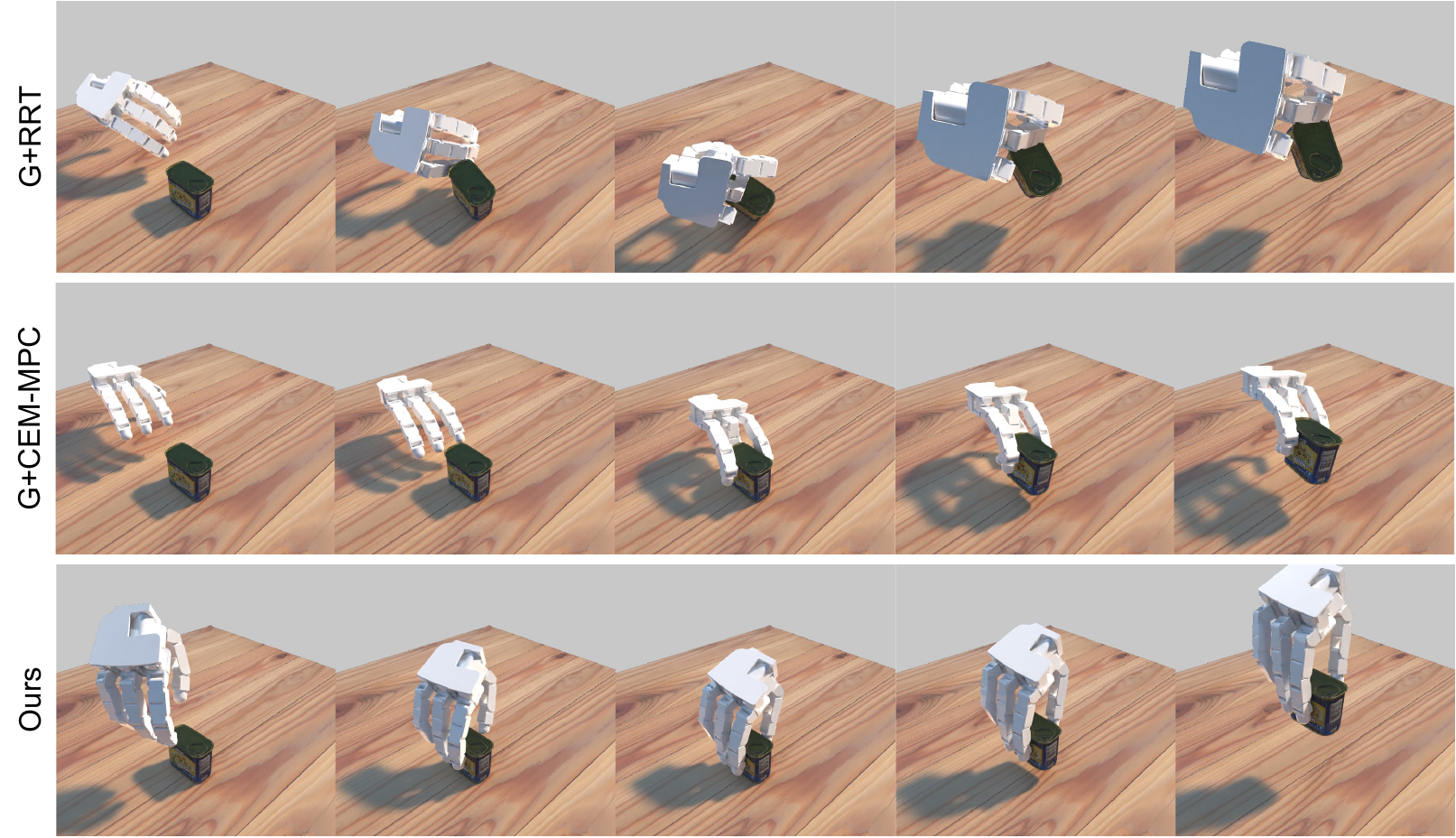}
  \vspace{-0.1in}
  \caption{\small
  Qualitative evaluation in the simulation. Because of human demonstrations, our CGF generates a more natural and reasonable trajectory, which is helpful for the sim-to-real transfer. G is short for GraspTTA.}
  \label{fig:compare}
  \vspace{-0.2in}
\end{figure*}

\vspace{-0.1in}
\subsection{Experimental Setting}
\label{subsec:setting}

\textbf{Datasets.} We utilize the DexYCB dataset~\cite{chao2021dexycb} to serve as human demonstrations. DexYCB contains 1,000 sequences of human grasping motions with 20 YCB objects~\cite{calli2015ycb}. We use 15 of them as training objects. We remove the handover process from the demonstrations, filter out all left-handed sequences and perform hand motion retargeting (Sec.~\ref{subsec:retargeting}) to translate the hand motion into Allegro robot motion.

\textbf{Baselines.} We mainly compare CGF to two-step methods, i.e., grasping synthesis followed by motion planning. We provide two grasping synthesis baselines: GraspTTA~\cite{jiang2021hand} and GraspIt~\cite{miller2004graspit}, where the former is also a generative model and the latter is based on searching.
For GraspTTA, we apply the same hand motion retargeting to obtain the Allegro hand joint positions.
Then, we utilize rapidly exploring random tree (RRT)~\cite{lavalle1998rapidly} and cross-entropy method (CEM)~\cite{rubinstein1999cross} with model predictive control (MPC)~\cite{maciejowski2002predictive} for planning the trajectory to reach the grasp pose, which are named GraspTTA+RRT and GraspTTA+CEM-MPC respectively.
For GraspIt, due to the high Degree-of-Freedom of the dexterous hand, we firstly leverage the large set of grasping poses from ContactGrasp~\cite{brahmbhatt2019contactgrasp} to construct a low-dimensional subspace via EigenGrasp~\cite{ciocarlie2007dexterous}. We then use GraspIt~\cite{miller2004graspit} and RRT for the grasping searching (including post-optimization) and motion planning respectively. We name it GraspIt+RRT. For the smoothness evaluation, we additionally perform linear interpolation between the beginning and ending joint positions generated by CGF and take it as a trivial baseline.

\textbf{Implementation Details.}
For training CGF, we sample $2,000$ points on the object mesh as the input object point clouds. During training, we employ the Adam optimizer with the learning rate $5e-4$, where the learning rate is reduced by half per 500 epochs. The batch size is 32. The training takes 1000 epochs in total. The dimension of the latent code $z$ is 256. CGF is a 4-layer MLP with channels (1281, 512, 256, 25) and internal ReLU activation. The loss weights are $\lambda_q = 1$, $\lambda_j = 10$, $\lambda_{KL} = 1e-3$ and $\lambda_c = 50$. We sample $10,000$ latent codes for CGF and output the trajectories. We then pick the successful trajectories in the simulator to evaluate in the real world and count the success rate. For simulation experiments, environments are built upon the SAPIEN~\cite{xiang2020sapien} simulator.

For GraspTTA, we generate 200 grasp poses for each object. For CEM-MPC, we set popsize $M=100$, time horizon $T=5$, number of elites $e=10$, and iterations $I=2$. For GraspIt, we search for valid grasp poses with 100 different random seeds and use the contact energy as the objective function. For motion planning with RRT, we set 10,000 nodes in the tree and set step size $\epsilon=0.01$ and probability of sampling $\beta=0.5$.

\vspace{-0.05in}
\subsection{Evaluation Metrics}

\textbf{Smoothness.} To measure the continuity of the generated grasping, we propose to compare the smoothness. We first normalize joint positions by making all joints start at 0 and end at 1 in all trajectories. Then we calculate discrete first-order and second-order gradients, i.e. velocity and acceleration. Smoothness is defined as the sum of the $\mathcal{L}1$ distances of position, velocity, and acceleration between neighboring frames.
Note that the linear interpolation is the upper bound with a joint position smoothness of $1.0$.

\textbf{Cost per Successful Trajectory in Simulator.}
While sampling more grasps with a generative model or more random configurations with a planning algorithm could increase the probability of finding a successful trajectory, the cost should not be neglected.
Thus, the average cost per successful trajectory is evaluated in the simulator for our method and baselines. The cost is defined as the number of environment steps for GraspTTA+CEM-MPC and our method or collision checks for GraspTTA+RRT, per successful trajectory. Additionally, wall-clock time is provided for a comprehensive evaluation.

\textbf{Success Rate in the Real World.} We also evaluate the success rate in the real world. Note the successful trajectories in the simulator do not guarantee success in the real world because of the sim-to-real gap. For our method and all the baselines, we collect 20 successful trajectories, deploy them in the real world and count the success rate. This metric reflects the sim-to-real transfer ability.

\begin{figure*}[t]
\centering
  \vspace{-0.1in}
  \includegraphics[width=0.9\linewidth]{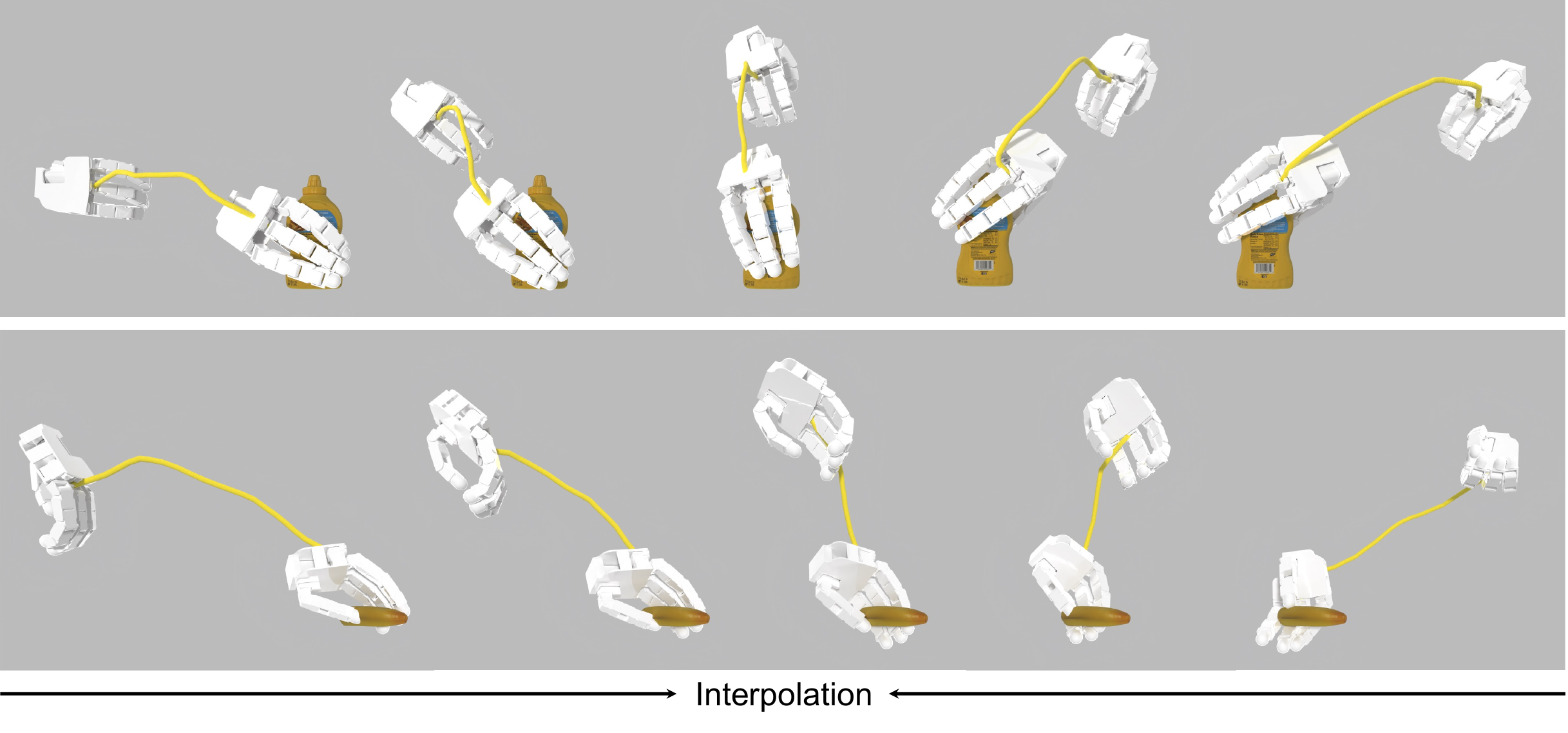}
  \vspace{-0.2in}
  \caption{\small
  Grasping interpolation. We show the first frame and the last frame of the grasping trajectory. Yellow lines indicate the trajectory of the palm joint. Our method produces diverse grasping and the interpolation between them is also plausible. To the best of our knowledge, this result on interpolating both robot hand grasping pose and trajectory has not been shown before.}
  \vspace{-0.2in}
  \label{fig:interpolation}
\end{figure*}

\begin{table}[t]
\tiny
\tablestyle{1pt}{1.05}
\centering
\vspace{-0.05in}
\begin{tabular}{l|ccc|ccc}
     &  \multicolumn{3}{c}{Smoothness - Joints $\downarrow$ } & \multicolumn{3}{c}{Smoothness - Cartesian $\downarrow$ } \\
    Method & Pos  & Vel (log)  & Acc (log)  & Pos  & Vel (log)  & Acc (log) \\
    \shline
    G + RRT & \textbf{3.40} & 3.43 & 6.47 & 9.71 & 3.95 & 6.99 \\
    G + CEM-MPC & 7.42 & 3.12 & 5.54 & 16.36 & 3.48 & 5.90 \\
    Ours & 10.42 & \textbf{2.06} & \textbf{3.51} & \textbf{6.69} & \textbf{1.99} & \textbf{3.47} \\
    \hline
    \demph{Linear Interpolation} & \demph{1.00} & \demph{-3.58} & \demph{-1.99} & \demph{1.96} & \demph{0.70} & \demph{1.38} \\
\end{tabular}
\vspace{-0.1in}
\caption{\small
For smoothness of joint positions and Cartesian coordinates, our CGF outperforms baselines by a large margin, which helps produce more natural trajectories and better control. G is short for GraspTTA.}
\vspace{-0.2in}
\label{tab:smoothness}
\end{table}

\vspace{-0.05in}
\subsection{Simulated Experimental Results}

\textbf{Smoothness Evaluation.} We compare the smoothness with three baselines and summarize the results in Tab.~\ref{tab:smoothness}. For better comparison, the smoothness for velocity and acceleration is in the log form. For both joint positions and Cartesian coordinates, our CGF outperforms baselines by a large margin. Note that for the joint position smoothness, our method does not show an advantage over the baseline. This is due to that humans tend to follow a natural curve rather than the shortest path. Nevertheless, the improvement in the smoothness of velocity and acceleration plays an important role in producing natural trajectories, suppressing vibration, and achieving high accuracy control~\cite{flash1985coordination, piazzi2000global}.

\begin{table}[t]
\tiny
\tablestyle{2pt}{1.05}
\centering
\begin{tabular}{l|cc|cc}
     & \multicolumn{2}{c}{Cost (log) / Succ. Traj. $\downarrow$} & \multicolumn{2}{c}{Time (s) / Succ. Traj. $\downarrow$} \\
     Method & Seen & Unseen & Seen & Unseen \\
    \shline
    G + RRT & 5.30$^*$ & 4.97$^*$ & 56.61 & 41.47 \\
    G + CEM-MPC & 5.85 & 5.58 & 146.78 & 123.48 \\
    Ours & \textbf{4.30} & \textbf{4.19} & \textbf{11.70} & \textbf{8.31} \\
    \hline
    \demph{Linear Interpolation} & \demph{-} & \demph{-} & \demph{-} & \demph{-} \\
\end{tabular}
\vspace{-0.05in}
\caption{\small
Success cost evaluation. Since our method only requires fewer simulation steps to test a trajectory, the average cost and time for a successful trajectory is much lower than baselines. Note that we calculate the amount of collision detection for RRT. G is short for GraspTTA.}
\label{tab:sim}
\vspace{-0.2in}
\end{table}

\textbf{Success Cost Evaluation.}
The average cost and time per successful trajectory are presented in Table~\ref{tab:sim}. The number of simulation steps and collision detections are shown in the log form for better comparison. Our method, which directly executes target joint positions generated from CGF, obtains a significantly lower average cost and time per successful trajectory compared to GraspTTA+RRT and GraspTTA+CEM-MPC.
The main reasons are: (i) Our method does not require excessive exploration like MPC-CEM or sampling numerous configurations like RRT; (ii) Our method produces more natural trajectories than two-step methods, yielding a higher probability for finding a successful trajectory.
We also evaluate on unseen objects, our method still surpasses the baselines with a slight decrease in metrics. This decrease in cost for unseen objects can be attributed to that the difficulty of finding a successful trajectory is largely determined by the object geometry. The linear interpolation baseline, which generates the smoothest trajectory, never grasps objects successfully in the simulator.

\textbf{Qualitative Evaluation.} We visualize the typical trajectories of the baselines and our method in Fig.~\ref{fig:compare}. Since RRT is only planning the reachable target joint positions, the trajectory may not be natural. And a small perturbation in the execution process may cause it to collide with objects. CEM-MPC is not a long-time
horizon method, which will make it fall into a local optimum quickly. As shown in Fig.~\ref{fig:compare}, the middle finger is not in the ideal position. However, our CGF, learning from the human demonstration, could generate a much more natural and smooth trajectory. This is helpful for the sim-to-real transfer, which we will discuss in the next section.

\textbf{Grasping Diversity.} The ability to generate diverse outputs is one of the motivations for using CVAE. We perform interpolation in the latent space and show the results in Fig.~\ref{fig:interpolation}. CGF produces diverse grasping and the interpolation between them is also plausible. We believe this is an interesting and potentially very useful property of our model. To the best of our knowledge, this result on interpolating both robot hand grasping pose and trajectory has not been shown before.

\begin{table}[t]
\tiny
\centering
\tablestyle{1.5pt}{1.05}
\begin{tabular}{l|cccccc}
     & \multicolumn{6}{c}{Success Rate (\%) $\uparrow$} \\
    Method & Banana & Cleanser & Meat Can & Soup Can & Bottle & Ball  \\
    \shline
    G + RRT & 10.0 & 15.0 & 10.0 & 5.0 & 5.0 & 10.0 \\
    G + CEM-MPC & 10.0 & 10.0 & 15.0 & 5.0 & 0.0  & 15.0 \\
    GI + RRT & 60.0 & 65.0 & 50.0 & 50.0 & 65.0 & 40.0 \\
    Ours & \textbf{70.0} & \textbf{85.0} & \textbf{70.0} & \textbf{80.0} & \textbf{85.0} & \textbf{65.0} \\
\end{tabular}
\vspace{-0.05in}
\caption{\small
Real-world experiments. For the successful trajectories in the simulation, our CGF has a higher success rate in the real world. G is short for GraspTTA and GI is short for GraspIt.}
\vspace{-0.15in}
\label{tab:real}
\end{table}

\begin{figure*}[t]
\centering
  \vspace{-0.1in}
  \includegraphics[width=0.95\linewidth]{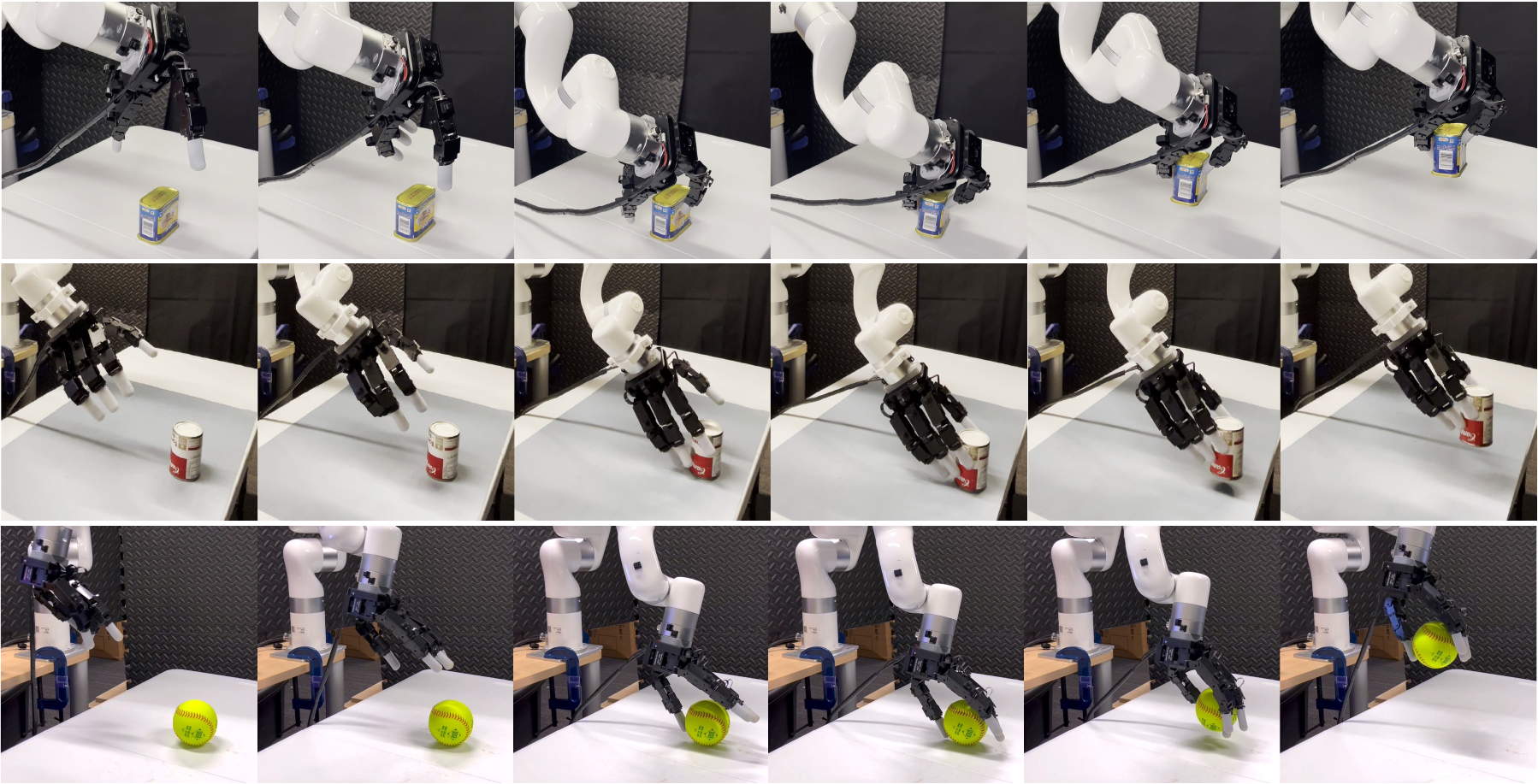}
  \vspace{-0.05in}
  \caption{\small
  Real-world results on \textit{Meat Can}, \textit{Soup Can} and \textit{Ball}. Our CGF successfully transfers the simulation trajectory to the real robot.}
  \vspace{-0.2in}
  \label{fig:real}
\end{figure*}

\vspace{-0.05in}
\subsection{Real-World Robot Experiments}
\textbf{Setup.} For the real-world robot experiments, we attached an Allegro hand on an X-Arm 6 robot. We select 6 real objects from YCB~\cite{calli2015ycb} which are \textit{Banana}, \textit{Bleach Cleanser}, \textit{Ball} \textit{Potted Meat Can}, \textit{Tomato Soup Can} and \textit{Mustard Bottle}, where the first 3 objects are unseen. The initial pose of the object is given and we use open-loop control for grasping.

\textbf{Results with a Real Robot Hand.}  We evaluate the sim-to-real transfer ability of the trajectories generated by our method and baselines. For each object and method, we collect 20 successful trajectories in the simulator and deploy them in the real world. We report the success rate in Tab.~\ref{tab:real}. Although all the trajectories succeeded in the simulator, our method has a much higher success rate in the real world than the baselines. We believe there are two main reasons leading to better sim-to-real transfer ability: (i) Learning from human demonstration can lead to more natural behavior trajectory; (ii) The use of implicit function also helps provide a continuous and more smooth trajectory. In contrast, the motion planning used in the 2-step procedure baselines often generates unnatural trajectories which reduces the success rate when deploying in the real world. Specifically, while GraspTTA~\cite{jiang2021hand} is able to generate grasp poses in the simulation, it does not ensure stable robotic grasping in the real world, and an unnatural trajectory approaching the grasp accumulates additional errors. On the other hand, GraspIt~\cite{miller2004graspit} is able to generate stable robot grasp pose in the real world, however, the 2-step procedure with it still performs worse than our method. We further provide a visualization of our successful trajectories in the real world in Fig.~\ref{fig:real}, which shows that our method could generate natural and human-like grasping.

\vspace{-0.1in}
\subsection{Ablation Study}
\vspace{-0.05in}

\begin{table}[t]
\tiny
\centering
\tablestyle{1.5pt}{1.05}
\begin{tabular}{l|ccc}
     & Mass 1x & Mass 2x  & Mass 3x \\
    \shline
    Velocity 1x & 5.66\% & 4.15\% & 7.10\%  \\
    Velocity 2x &  14.95\% & 7.73\%  & 2.10\%\\
    Velocity 3x & 4.29\% & 1.65\%  & 16.77\%  \\
\end{tabular}
\vspace{-0.1in}
\caption{\small
Ablation study on the PD control with feedforward. We report improvements on the success rate of the PD control with feedforward over the default PD control with different velocities and masses.}
\vspace{-0.15in}
\label{tab:augpd}
\end{table}

To demonstrate the advantages of continuous grasping, we ablate the PD control with feedforward (Sec.~\ref{sec:pd-control}) which relies on the derivatives of CGF. Note that baselines can not directly utilize PD control with feedforward since they are not continuous. We compare the performance of the PD control with feedforward to the default PD control in the simulator with different velocities and masses. Specifically, we set the speed to 1x, 2x and 3x and the mass of the robot to 1x, 2x and 3x. We combine them in pairs and report the improvements on the success rate in Tab.~\ref{tab:augpd}. We show that the PD control with feedforward has improvements in all various cases, which proves its robustness.

\begin{table}[t]
    \centering
    \tablestyle{1.5pt}{1.05}
    \begin{tabular}{c|cc|cc}
         & \multicolumn{2}{c}{Cost/S.T. - Noisy Points $\downarrow$} & \multicolumn{2}{c}{Cost/S.T. - Noisy Pose $\downarrow$} \\
        $s$ & Seen & Unseen & Seen & Unseen \\
        \shline
        0 & 4.30 & 4.19 & 4.30 & 4.19 \\
        0.001 & 4.28 & 4.26 & 4.31 & 4.38 \\
        0.01 & 4.32 & 4.38 & 4.38 & 4.56 \\
        0.1 & 4.39 & 4.67 & 4.48 & 5.64 \\
        1 & 4.61 & 4.89 & - & - \\
    \end{tabular}
    \vspace{-0.1in}
    \caption{\small
    Analysis Study for Noises. Our method is robust against noises on both point cloud and pose.
    }
    \vspace{-0.15in}
\label{tab:noise}
\end{table}

In addition, we analyze the robustness of CGF by adding noises on the input point cloud and the object pose. For the object point cloud $\mathcal{P}^{o} \in \mathbb{R}^{N \times 3}$, we add $N$ independently sampled Gaussian noises $n \sim \mathcal{N}(0, \Sigma)$, where $\Sigma = \text{diag}(\sigma)$ is the diagonal covariance matrix. The noise scale was set with $\sigma = s \left( \max_i{\mathcal{P}^{o}_i} - \min_i{\mathcal{P}^{o}_i} \right)$, with $s$ as the parameter controlling the noise scale and $\mathcal{P}^{o}_i$ is the $i$-th point in $\mathcal{P}^{o}$. For the object pose $T \in \mathbb{S E}(3)$, we also add Gaussian noises $n \sim \mathcal{N}(0, \Sigma)$ to its translation part, with the same distribution as that for the point clouds. The results in Tab.~\ref{tab:noise} show that our method is robust against both point cloud and pose noise, despite the higher cost of finding successful trajectories resulting from increased noise.

\vspace{-0.05in}
\section{Conclusion}
\label{sec:discussion}

We propose a novel implicit function named Continuous Grasping Function to generate smooth and dense grasping trajectories. CGF is learned in the framework of a Conditional Variational AutoEncoder using 3D human demonstrations. During inference, we sample various grasping plans in the simulator and deploy the successful ones to the real robot. By training on diverse human-object data, our method allows generalization to manipulate multiple objects. Compared to previous planning algorithms, CGF is more efficient and has a better sim-to-real generalization ability.

{\small
    \bibliographystyle{IEEEtran}
    \bibliography{IEEEabrv,reference_simplified.bib}
}
\end{document}